\documentclass{article}
\usepackage{spconf}
\usepackage{cite}
\usepackage{amsmath,amssymb,amsfonts}
\usepackage{graphicx}
\usepackage{textcomp}
\usepackage{xcolor}
\usepackage{multirow}
\usepackage{arydshln}
\usepackage{todonotes}
\usepackage{etoolbox}
\usepackage{setspace}
\usepackage[bottom]{footmisc}
\usepackage{placeins}
\usepackage{wrapfig}
\usepackage{color, colortbl}
\usepackage[export]{adjustbox}
\usepackage{amsmath}
\usepackage{algpseudocode}
\usepackage[ruled,vlined]{algorithm2e}
\usepackage{tikz}
\usepackage{relsize}
\usepackage{mathtools, nccmath}
\usepackage{url}
\usepackage{float}
\usepackage{siunitx}
\graphicspath{ {./images/} }

\newcommand{\CC}[1]{\cellcolor{white}}

\begin{document}
\title{On the impact of using X-ray energy response imagery for object detection via Convolutional Neural Networks}

%
\name{Neelanjan Bhowmik$^1$, Yona Falinie A. Gaus$^1$, Toby P. Breckon$^{1,2}$}
\address{Department of \{Computer Science$^1$ $|$ Engineering$^2$\}, Durham University, UK}

%
\maketitle

\begin{abstract}
Automatic detection of prohibited items within complex and cluttered X-ray security imagery is essential to maintaining transport security, where prior work on automatic prohibited item detection focus primarily on pseudo-colour ({\it rgb}) X-ray imagery. In this work we study the impact of variant X-ray imagery, i.e., X-ray energy response ({\it high, low}) and effective-$z$ compared to {\it rgb}, via the use of deep Convolutional Neural Networks (CNN) for the joint object detection and segmentation task posed within X-ray baggage security screening. We evaluate state-of-the-art CNN architectures (Mask R-CNN, YOLACT, CARAFE and Cascade Mask R-CNN) to explore the transferability of models trained with such `raw' variant imagery between the varying X-ray security scanners that exhibits differing imaging geometries, image resolutions and material colour profiles. Overall, we observe maximal detection performance using CARAFE, attributable to training using combination of {\it rgb, high, low}, and effective-$z$ X-ray imagery, obtaining $0.7$ mean Average Precision (mAP) for a six class object detection problem. Our results also exhibit a remarkable degree of generalisation capability in terms of cross-scanner transferability (AP: $0.835/0.611$) for a one class object detection problem by combining {\it rgb, high, low}, and effective-$z$ imagery.
\end{abstract}
\begin{keywords}
x-ray imagery, deep convolutional neural network, object detection, transferability
\end{keywords}

    \vspace{-0.3cm}
    \section{Introduction} \label{s:intro}
\begin{figure}[htb!]
\centering
\includegraphics[width=\linewidth]{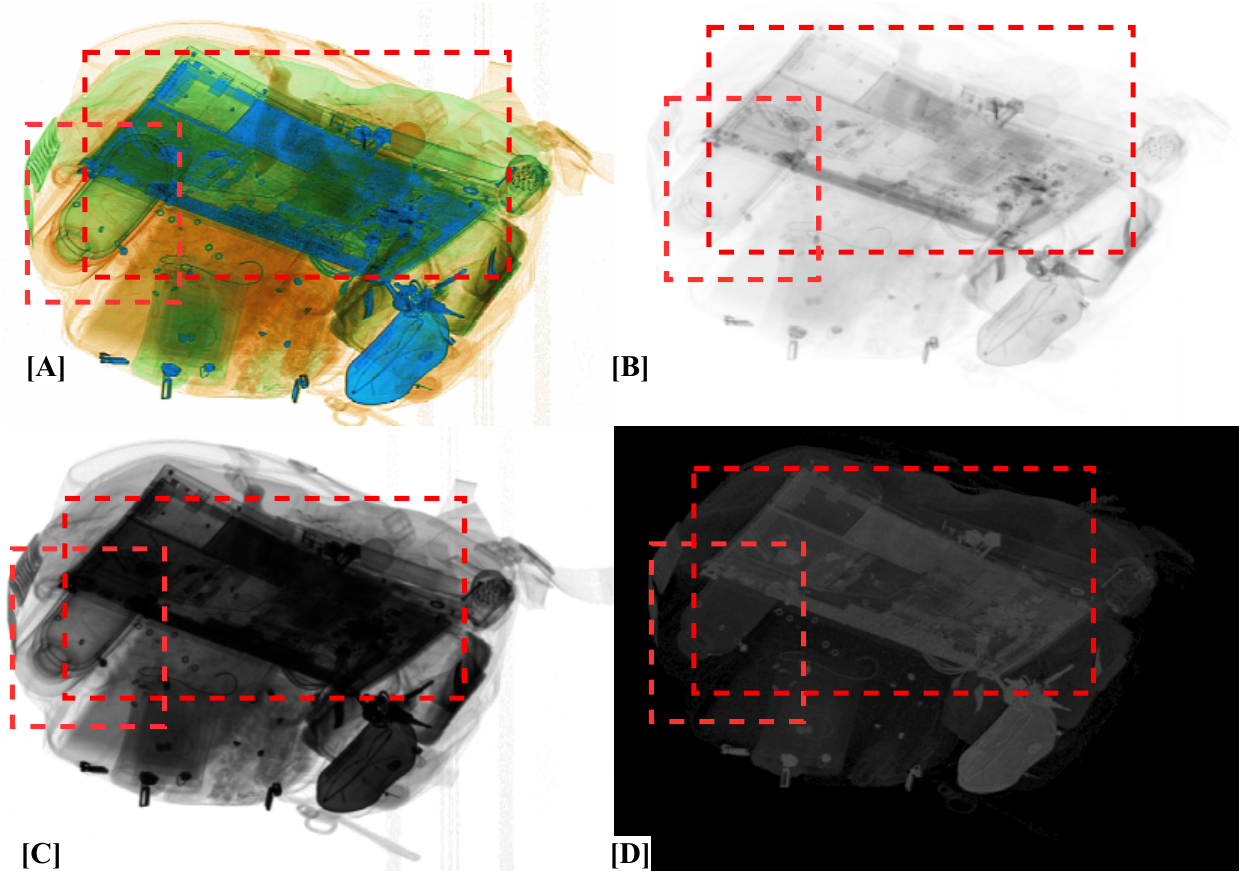}
\vspace{-0.8cm}
\caption{Exemplar {\it rgb} (A), {\it high} (B), {\it low} (C), and effective-$z$ (D) X-ray imagery from {\it deei6} dataset containing target classes in bounding boxes.}
\label{fig:ex_fchlz}
\vspace{-0.8cm}
\end{figure}

\vspace{-0.2cm}
X-ray security screening plays a pivotal role in aviation security. However, manual inspection of potentially prohibited items is challenging due to the clutter and occlusion present within X-ray scanned baggage. 
A modern X-ray security scanner makes use of multiple X-ray energy levels in order to facilitate effective materials discrimination \cite{singh2003explosives}. Subsequently, a dual-energy X-ray scanner imagery consists of two intensity images acquired at two discrete energy levels ({\it low} and {\it high}), facilitating the recovery of material properties (effective atomic number, effective-$z$). The information is fused with the help of a colour transfer function into a single pseudo-colour X-ray image (Figure \ref{fig:ex_fchlz}A) to facilitate the interpretation of the baggage contents \cite{turcsany13xray}.

\begin{figure*}[htb!]
\centering
\includegraphics[width=17cm]{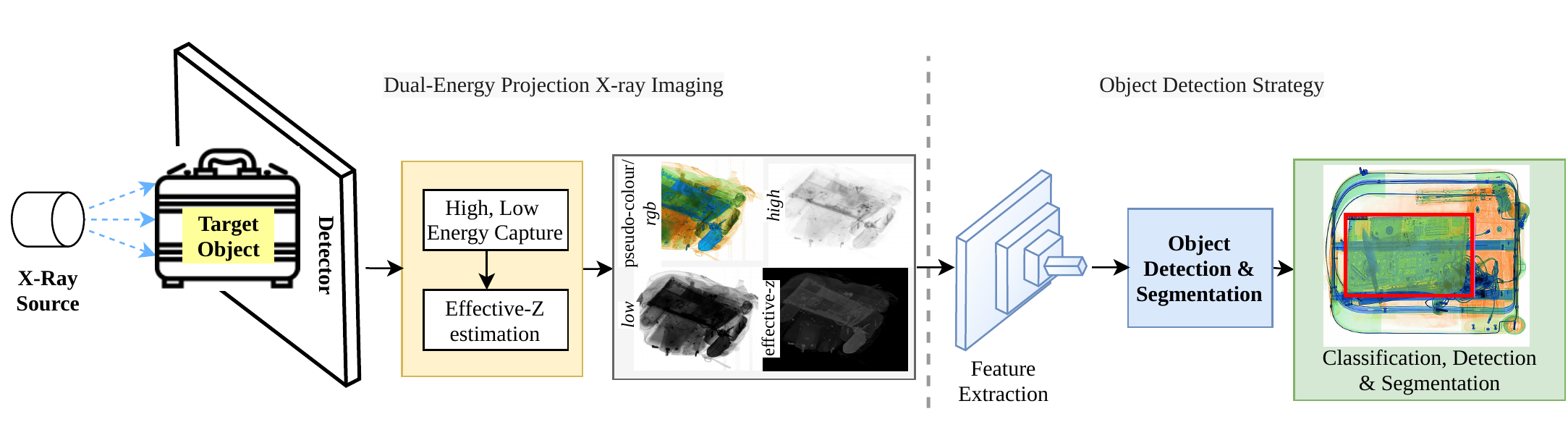}
\vspace{-0.8cm}
\caption{Schematic of X-ray imaging followed by CNN architecture for object detection in complex X-ray security imagery.}
\label{fig:arch}
\vspace{-0.5cm}
\end{figure*}

The advancement of deep Convolutional Neural Networks (CNN) has brought new insight to the automation of this X-ray imagery screening task \cite{akccay2016transfer,Akcay2018Xray,gaus2019evaluation} where the primary task is both to localise and classify the prohibited items. Prior works \cite{Akcay2018Xray,gaus2019evaluating} are concentrated on the shaped-based detection of prohibited items achieving both high detection performance with low false positive. The work of \cite{akccay2016transfer} uses a pre-trained GoogleNet model for classification task in X-ray baggage scans for detecting potentially prohibited items. Subsequently, the work of \cite{Akcay2018Xray} compares contemporary region-based and single forward-pass based CNN architectures (Faster R-CNN \cite{Ren2015fasterr-cnn}, YoloV2 \cite{Redmon2016yolo}) achieving $0.88$ and $0.97$ mean Average Precision (mAP) over six class and two class X-ray baggage security object detection problems respectively. Following these works, \cite{gaus2019evaluation} proposes a dual-stage CNN architecture for anomaly detection in a six class problem. Semi-supervised adversarial learning is used in the works of \cite{akcay2018ganomaly,akccay2019skip} for prohibited item detection. The availability of the large-scale X-ray baggage datasets (SIXray \cite{Miao2019SIXray}, and OPIXray \cite{wei2020occluded}) has provided further insight into the transferability and generalisation abilities of the CNN architectures \cite{gaus2019evaluating} across varying X-ray security scanners which all exhibit varying characteristics in terms of projection geometry common resolution and pseudo-colour mapping. While most prior work \cite{Akcay2018Xray,gaus2019evaluation,gaus2019evaluating} process each view of multi-view X-ray scanners independently, \cite{isaac20multiview} utilise the corresponding information between the views for a detection task achieving $0.91$ mAP.

Almost all of the prior work discussed here only use pseudo-colour/false colour ({\it rgb}) X-ray imagery that is itself generated from the `raw' {\it high/low/}effective-$z$ imagery obtained from the scanner. By contrast, in this study we consider the impact of using this `raw' imagery (Figure \ref{fig:ex_fchlz}(B)$\rightarrow$(D)) directly for the purposes of prohibited object detection. The objective of using two energy levels ({\it high} and {\it low}) for object detection task is to obtain both the density and atomic number $Z$ (effective-$z$) of the scanned materials \cite{rebuffel2007dual}, as the intensity values in the energy response may encode very valuable material information, which is not as readily identifiable within the pseudo-colour X-ray imagery. 

Against this background, this paper introduces the following novel contributions: (a) an experimental evaluation of dual-energy X-ray imagery for joint object detection and segmentation task, via use of characteristically diverse end-to-end CNN architectures \cite{he2017maskrcnn,Bolya19:yolact,wang19:carafe,cai19:cascade}, (b) an investigation into the inter-scanner transferability of such CNN models, trained on dual-energy X-ray imagery, in terms of their generalisation across varying X-ray scanner characteristics.

    \vspace{-0.4cm}
\section{Proposed Approach} \label{s:proposal}
\vspace{-0.2cm}

In this study, we present dual-energy X-ray imaging technique (Figure \ref{fig:arch}, left) in Section \ref{ssec:xray_formation} and followed by object detection and segmentation strategies (Figure \ref{fig:arch}, right) in Section \ref{ssec:seg}. 

\vspace{-0.4cm}
\subsection{Dual-Energy Projection X-ray Imaging} \label{ssec:xray_formation}
\vspace{-0.2cm}
The primary components of X-ray security scanner system are composed of an X-ray source emitter and detector (Figure \ref{fig:arch}, left). X-rays are emitted with photon energy ranging up to 150kV \cite{gilardoni_scanner640} from a X-ray source. Generally, the X-ray images are constructed by attenuating the signal on the material as the target object proceeds through the scanner tunnel, defined as $I(E)=I_0e^{-\mu t}$, where $I(E)$ is the captured intensity as a function of the thickness $t$, the emitted intensity $I_0$ and the absorption coefficient $\mu$. The absorption coefficient is defined by $\mu = \alpha(Z,E)\rho$, where $Z$ is the atomic number, $E$ is the energy, $\rho$ is the density, and $\alpha(Z,E)$ corresponds to the mass attenuation coefficient in terms of $Z$ and $E$ \cite{mery2020x}.  

In the dual-energy source X-ray imaging, two intensity responses captured at two different energy levels, {\it low} and {\it high} ($E=\{l,h\}$) and are subsequently combined to construct {\it low} and {\it high} energy response images (Figure \ref{fig:arch}, left). Given the Compton scatter coefficient ($\mu_c$) and the photoelectric absorption coefficient ($\mu_p$) \cite{Mouton2015ARO}, material identification (approximate atomic number, effective-$z$; $Z\_eff$) can be calculated as:
\begin{equation}
    Z\_eff = K^{'}(\frac{\mu_p}{\mu_c})^\frac{1}{n}
\end{equation} 
where $K^{'}$ and $n$ are constant \cite{Mouton2015ARO}. 
In this work, we evaluate the use of the pseudo-colour ({\it rgb}), dual-energy response ({\it h, l}) and effective-$z$ (Figure \ref{fig:ex_fchlz}) as alternative inputs imagery for CNN-based object detection.

\vspace{-0.3cm}
\subsection{Object Detection and Segmentation Strategy} \label{ssec:seg}
\vspace{-0.2cm}
We consider four contemporary CNN architectures of differing characteristics, spanning both single stage and multi stage detection approaches, and explore their applicability for prohibited item detection within varying configurations of dual-energy X-ray imagery inputs. \\
{\bf Mask R-CNN \cite{he2017maskrcnn}} is a two-stage detector for object instance segmentation, developed on top of Faster R-CNN \cite{Ren2015fasterr-cnn}. 
Mask R-CNN \cite{he2017maskrcnn} uses the Faster R-CNN \cite{Ren2015fasterr-cnn} architecture for feature extraction, Region Proposal Network (RPN), and followed by region of interest alignment (RoIAlign) via bilinear boundary interpolation to produce higher resolution feature map boundaries suitable for input into a secondary classifier. The output from the RoIAlign layer is subsequently fed into a series of segmentation processing layers (mask head), that generate an additional image mask indicating pixel membership of a given detected object. \\
{\bf YOLACT \cite{Bolya19:yolact}} is an one-stage detector, based on RetinaNet \cite{lin2017retinanet}, that directly predicts boxes without a separate region proposal step. YOLACT \cite{Bolya19:yolact} generates a set of prototype masks, linear combination coefficients for each predicted instance, and associated bounding boxes. It combines the prototype masks using the corresponding predicted mask coefficients followed by cropping with a predicted bounding box to generate the final output. \\
{\bf CARAFE \cite{wang19:carafe}} is a two-stage architecture, which proposes effective feature up-sampling operators and integrates it into Feature Pyramid Network to boost the performance. For instance segmentation, a feature map, which represents the object shape accurately, is used to predict the final instance segmentation result. \\
{\bf Cascade Mask R-CNN \cite{cai19:cascade}}, a multi-stage detector, is a hybrid of Cascade R-CNN and Mask R-CNN \cite{he2017maskrcnn}. Similar to Mask R-CNN \cite{he2017maskrcnn}, each stage has a segmentation mask branch, a label prediction branch, and a bounding box detector branch. The current stage will accept RPN or the bounding box returned by the previous stage as an input. The second stage increases localisation performance accuracy, and subsequently, it further refines the output. This is repeated over multiple stages with increasingly refined criteria for discarding low-quality proposals from the previous stage such that it predicts precise bounding boxes and masks at the final stage.

In this study, we compare these four CNN architectures for object detection (Figure \ref{fig:arch}, right) using combination of different variants dual-energy X-ray imagery (Section \ref{ssec:xray_formation}). To assess the impact of dual-energy X-ray imagery variants on object detection we first use {\it rgb}, {\it high} ($h$), {\it low} ($l$), and effective-$z$ ($z$) imagery individually. Secondly, $h$, $l$ and $z$ are combined as three channels ({\it hlz}) images. Thirdly, we combine {\it rgb, high, low}, and effective-$z$ imagery for joint object detection and segmentation task.

Within the X-ray imagery security domain, imagery may be sourced using varying scanners \cite{gilardoni_scanner640,smithsdetection_scanner,rapiscan_scanner_620}, which have different X-ray energy spectra, spatial resolution and material colour profiles. In prior work \cite{Caldwell2017transfer, gaus2019evaluating} on transferability and generalisation ability, \cite{Caldwell2017transfer} focuses on transfer learning between cargo parcel scanning (different scanner equipment due to the differences in scale). The work of \cite{gaus2019evaluating} shows cross-scanner transferability of CNN architectures (using {\it rgb} X-ray imagery) in terms of their generalisation across varying X-ray scanner characteristics.

In this study, we further evaluate the effectiveness of using variants of dual-energy X-ray imagery (Section \ref{ssec:xray_formation}) on generalisation capabilities of the CNN architectures.
    \vspace{-0.8cm}
\section{Evaluation} \label{s:eval}
\vspace{-0.3cm}
We focus on three datasets that are sourced from different X-ray scanners \cite{gilardoni_scanner640,smithsdetection_scanner,rapiscan_scanner_620}. The {\it deei6} is created from a Gilardoni X-ray scanner \cite{gilardoni_scanner640}, and consists of {\it rgb}, {\it high, low}, and effective-$z$ imagery. The other two datasets, {\it dbs\_laptop} and {\it dbr\_laptop}, are generated by a Smith Detection \cite{smithsdetection_scanner} and Rapiscan X-ray scanner \cite{rapiscan_scanner_620} respectively and consist of {\it rgb} X-ray imagery. 
The four CNN architectures (Section \ref{ssec:seg}) are trained using {\it rgb} and combinations of {\it rgb, high, low}, and effective-$z$ X-ray imagery from {\it deei6} dataset. Subsequently, we evaluate the model performance on {\it rgb} X-ray imagery of {\it dbs\_laptop} and {\it dbr\_laptop} datasets. \\
{\bf deei6:} Our dataset (Durham Electrical and Electronics Items) is constructed using a dual-energy Gilardoni FEP ME 640 AMX scanner \cite{gilardoni_scanner640} with associated pseudo-colour materials mapping. This dataset is composed of six-classes of consumer electronics, electrical and other items: \{{\it bottle, hairdryer, iron, toaster, phone-tablet, laptop}\}, totalling $7,022$ images (70:30 data split for experiments). We also access the {\it high, low}, and effective-$z$ imagery to construct {\it deei6$_{rgb}$},  {\it deei6$_{h}$}, {\it deei6$_{l}$} and {\it deei6$_{z}$} imagery as depicted in Figure \ref{fig:ex_fchlz}. \\
To investigate the generalisation capabilities of the CNN architectures, we also use the following two datasets: \\
{\bf dbs\_laptop:} comprises $488$ {\it laptop} class  {\it rgb} X-ray image examples (with associated pseudo-colour materials mapping), which is sourced from a Smith Detection X-ray scanner \cite{smithsdetection_scanner}. \\
{\bf dbr\_laptop:}  comprises $107$ {\it laptop} class X-ray {\it rgb} image examples (with associated pseudo-colour materials mapping). This dataset is sourced from Rapiscan 620DV X-ray scanner \cite{rapiscan_scanner_620}.

The CNN architectures (Section \ref{ssec:seg}) are implemented using MMDetection framework \cite{mmdetection}. 
Through the transfer learning paradigm, training (using X-ray imagery variants) of all CNN architectures (Section \ref{ssec:seg}) are initialised with ImageNet \cite{deng2009imagenet} pretrained weights (which originate from training on colour RGB imagery). Our CNN architectures are trained using ResNet$_{50}$ \cite{He15:ResNet} backbone with following training configuration: backpropagation optimisation performed via Stochastic Gradient Descent, initial learning rate of $\num{2.5e-4}$ with decay by a factor of $10$ at 7$^{th}$ epoch, and a batch size of $4$. 
The model performance is evaluated by MS-COCO metrics \cite{lin2014coco} (IoU of $0.50:.05:0.95$), using Average Precision (AP) for class-wise and mAP for overall performance.

\vspace{-0.4cm}
\subsection{Impact of Dual-energy X-ray Imagery} \label{ssc:baseline}
\vspace{-0.2cm}
In the first set of experiments (Table \ref{Table:mAP_impact}), exemplar items in X-ray security imagery are detected using the CNN architectures set out in Section \ref{ssec:seg}. We use variants of dual-energy X-ray imagery of the {\it deei6} dataset for training and evaluation denoted as {\it deei6$_{x}$} for $x = \{rgb, h, l, z, hlz\}$. The highlighted mAP signifies the maximal results obtained for overall performance. At first, the CNN architectures are trained and evaluated on {\it rgb} X-ray imagery (Table \ref{Table:mAP_impact}, {\it rgb}), in line with \cite{Akcay2018Xray,Miao2019SIXray,gaus2019evaluation}. The best performance is achieved by Cascade Mask R-CNN (CM RCNN) \cite{cai19:cascade} producing maximal mAP ($0.693$) and outperforming other three CNN architectures. When we train CNN architectures using {\it high}, {\it low}, and effective-$z$ imagery individually and together as three channels ({\it hlz}), the overall performance (Table \ref{Table:mAP_impact}) does not improve compared to {\it rgb} imagery. The lowest performing training set is {\it deei6$_{z}$} imagery achieving only $0.627$ of mAP (with Cascade Mask R-CNN \cite{cai19:cascade}).  It is possibly due to the lack of contrast in the pixel intensity in effective-$z$ imagery where the target objects appear similar to the background, leading to inferior detection performance. The impact of dual-energy X-ray imagery can be observed while combining {\it rgb, high, low} and effective-$z$ ({\it deei6$_{rgb,hlz}$}) together. The maximal mAP of $0.7$ (Table \ref{Table:mAP_impact}, {\it rgb,hlz}) is achieved by CARAFE \cite{wang19:carafe} marginally outperforming {\it rgb} imagery (mAP: $0.693$). 
Although YOLACT \cite{Bolya19:yolact} is the simplest architecture ($34.76$ million parameters), it outperforms Mask R-CNN \cite{he2017maskrcnn} while training using {\it rgb,hlz} X-ray imagery (mAP: $0.686$ vs $0.680$).

\begin{figure}[htb!]
\vspace{-0.4cm}
\centering
\includegraphics[width=\linewidth]{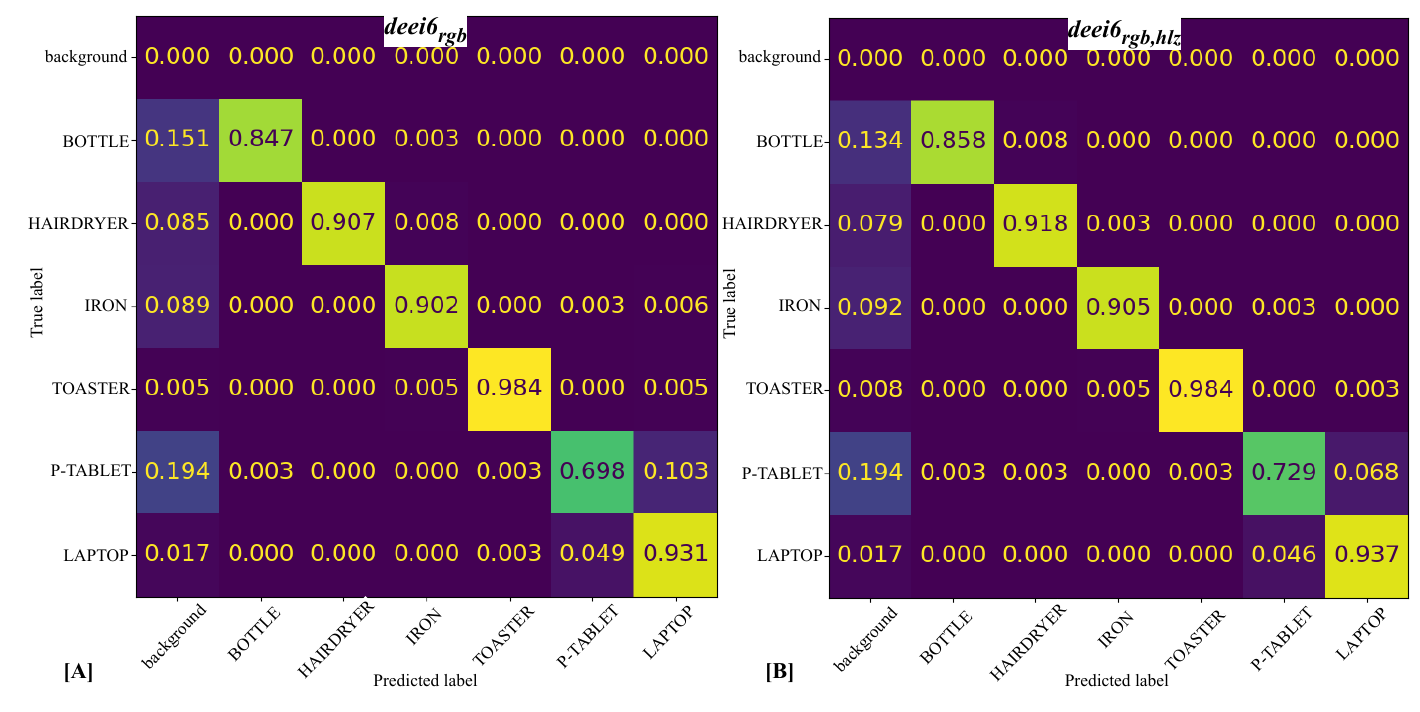}
\vspace{-0.9cm}
\caption{Confusion Matrix of the CARAFE \cite{wang19:carafe} trained on {\it rgb} (A) and combination of \{{\it rgb,hlz}\} (B) X-ray imagery.}
\label{fig:conf}
\vspace{-0.6cm}
\end{figure}

In the confusion matrices (Figure \ref{fig:conf}) of CARAFE \cite{wang19:carafe}, we observe strong true positive (diagonal) and low false positive (off-diagonal) occurrence. The advantage of combining {\it rgb, high, low}) and effective-$z$ can be seen in the class {\it phone-tablet} ($0.698$ to $0.729$, Figure \ref{fig:conf}(A)$\rightarrow$(B)), with improvement of confidence in localising small objects within cluttered X-ray security imagery. 
\vspace{-0.2cm}

\newcolumntype{g}{>{\columncolor{blue!5}}l}
\begin{table}[htb!]
\centering
\tiny
\begin{tabular}{lgggggggg}
\hline
\rowcolor{white}
& Model & Bottle & Hairdryer & Iron & Toaster & P-tablet & Laptop & mAP  \\  \hline
\multirow{4}{*}{\rotatebox[origin=c]{90}{\it deei6$_{rgb}$}} & M RCNN & 0.633 & 0.651 & 0.688 & 0.793 & 0.550 & 0.747 & 0.677 \\ 
& \CC{0}YOLACT  & \CC{0}0.646 & \CC{0}0.596 & \CC{0}0.672 & \CC{0}0.784 & \CC{0}0.540 & \CC{0}0.770 & \CC{0}0.668 \\ 
& CARAFE & 0.637 & 0.638 & 0.692 & 0.788 & 0.543 & 0.770  & 0.678 \\ 
& \CC{0}CM RCNN & \CC{0}0.650 & \CC{0}0.659 & \CC{0}0.708 & \CC{0}0.801 & \CC{0}0.560 & \CC{0}0.781 & \CC{0}{\bf 0.693} \\ \hline

\multirow{4}{*}{\rotatebox[origin=c]{90}{\it deei6$_{h}$}} & M RCNN & 0.607 & 0.615 & 0.665 & 0.761 & 0.521 & 0.745 & 0.652 \\
& \CC{0}YOLACT & \CC{0}0.641 & \CC{0}0.597 & \CC{0}0.649 & \CC{0}0.756 & \CC{0}0.533 & \CC{0}0.765 & \CC{0}0.657 \\ 
& CARAFE & 0.631 & 0.624 & 0.676 & 0.754 & 0.522 & 0.757 & 0.661 \\ 
& \CC{0}CM RCNN & \CC{0}0.632 & \CC{0}0.638 & \CC{0}0.687 & \CC{0}0.782 & \CC{0}0.539 & \CC{0}0.783 & \CC{0}{\bf 0.677} \\ \hline 

\multirow{4}{*}{\rotatebox[origin=c]{90}{\it deei6$_{l}$}} & M RCNN & 0.597 & 0.605 & 0.670 & 0.779 & 0.520 & 0.749 & 0.653 \\ 
& \CC{0}YOLACT & \CC{0}0.619 & \CC{0}0.576 & \CC{0}0.659 & \CC{0}0.771 & \CC{0}0.520 & \CC{0}0.760 & \CC{0}0.651  \\ 
& CARAFE & 0.632 & 0.606 & 0.662 & 0.777 & 0.530 & 0.768 & 0.662 \\ 
& \CC{0}CM RCNN & \CC{0}0.641 & \CC{0}0.627 & \CC{0}0.677 & \CC{0}0.784 & \CC{0}0.541 & \CC{0}0.778 & \CC{0}{\bf 0.674} \\ \hline

\multirow{4}{*}{\rotatebox[origin=c]{90}{\it deei6$_{z}$}} & M RCNN & 0.543 & 0.521 & 0.629 & 0.798 & 0.489 & 0.716 & 0.616  \\ 
& \CC{0}YOLACT & \CC{0}0.548 & \CC{0}0.395 & \CC{0}0.597 & \CC{0}0.783 & \CC{0}0.477 & \CC{0}0.737 & \CC{0}0.589 \\ 
& CARAFE & 0.550 & 0.492 & 0.629 & 0.786 & 0.522 & 0.718 & 0.616 \\ 
& \CC{0}CM RCNN & \CC{0}0.560 & \CC{0}0.516 & \CC{0}0.634 & \CC{0}0.796 & \CC{0}0.507 & \CC{0}0.749 & \CC{0}{\bf 0.627} \\ \hline

\multirow{4}{*}{\rotatebox[origin=c]{90}{\it deei6$_{hlz}$}} & M RCNN & 0.613 & 0.617 & 0.667 & 0.789 & 0.535 & 0.742 & 0.660  \\ 
& \CC{0}YOLACT & \CC{0}0.615 & \CC{0}0.575 & \CC{0}0.644 & \CC{0}0.757 & \CC{0}0.525 & \CC{0}0.756 & \CC{0}0.645 \\ 
& CARAFE & 0.639 & 0.611 & 0.673 & 0.791 & 0.557 & 0.765 & 0.673 \\ 
& \CC{0}CM RCNN & \CC{0}0.632 & \CC{0}0.630 & \CC{0}0.689 & \CC{0}0.802 & \CC{0}0.541 & \CC{0}0.775 & \CC{0}{\bf 0.678} \\ \hline

\multirow{4}{*}{\rotatebox[origin=c]{90}{\shortstack[l]{{\it deei6} \\ {\it $_{rgb,hlz}$}}}} & M RCNN & 0.644 & 0.633 & 0.682 & 0.799 & 0.543 & 0.779 & 0.680 \\
& \CC{0}YOLACT & \CC{0}0.670 & \CC{0}0.625 & \CC{0}0.676 & \CC{0}0.796 & \CC{0}0.560 & \CC{0}0.791 & \CC{0}0.686 \\ 
& CARAFE & 0.676 & 0.653 & 0.690 & 0.808 & 0.580 & 0.792 & \underline{\bf 0.700}  \\
& \CC{0}CM RCNN & \CC{0}0.667 & \CC{0}0.663 & \CC{0}0.696 & \CC{0}0.806 & \CC{0}0.552 & \CC{0}0.798 & \CC{0}0.697 \\ \hline

\end{tabular}
\vspace{-0.3cm}
\caption{Object detection results of CNN architectures using different X-ray imagery from the {\it deei6} dataset.}
\label{Table:mAP_impact}
\vspace{-0.4cm}
\end{table}

\vspace{-0.2cm}
\subsection{Cross-scanner Transferability} 
\label{ssc:gen}
\begin{figure}[htb!]
\vspace{-0.6cm}
\centering
\includegraphics[width=7.5cm]{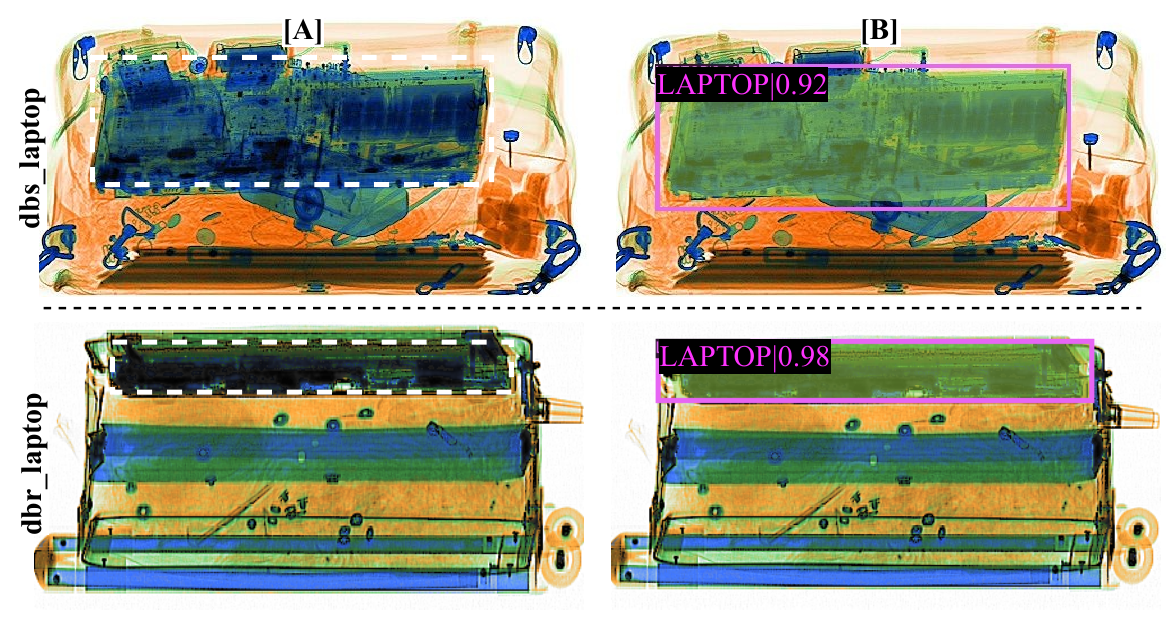}
\vspace{-0.6cm}
\caption{Detection examples from {\it dbs$\_$laptop} and {\it dbr$\_$laptop} using CARAFE \cite{wang19:carafe} trained on {\it rgb} (A) and \{{\it rgb,hlz}\} (B) X-ray imagery from {\it deei6} dataset. White dashed box in (A) fails to detect the target.}
\label{fig:det_ex1}
\vspace{-0.2cm}
\end{figure}

\vspace{-0.2cm}
In this set of experiments (Table \ref{Table:mAP_generalise}), we assess the CNN architecture performance across the X-ray imagery ({\it dbs$\_$laptop} and {\it dbr$\_$laptop}) from different scanner sources \cite{smithsdetection_scanner,rapiscan_scanner_620}. The CNN architectures are trained using variants of dual-energy X-ray imagery of {\it deei6} dataset but evaluated on a test set of only $rgb$ pseudo-colour imagery ({\it dbs$\_$laptop} and {\it dbr$\_$laptop}). The positive impact of combining \{{\it rgb,hlz}\} X-ray imagery is evident with all four CNN architectures (Table \ref{Table:mAP_generalise}). For {\it dbs$\_$laptop}, CARAFE \cite{wang19:carafe} produces the best performance (AP: $0.835$, Table \ref{Table:mAP_generalise}, lower) when trained using combination of {\it rgb, high, low} and effective-$z$ X-ray imagery, significantly outperforming {\it rgb} X-ray imagery (AP: $0.763$, Table \ref{Table:mAP_generalise}, upper). Similar significant performance improvement is noticeable on {\it dbr$\_$laptop} dataset with CARAFE \cite{wang19:carafe} achieving the highest AP of $0.611$ (Table \ref{Table:mAP_generalise}, lower). A plausible explanation for the performance improvement is that the variation in X-ray imagery by combining {\it rgb, high, low} and effective-$z$ imagery during training, leads the CNN architectures to learn meaningful image features, which alleviates to achieve a higher degree of model generalisation in object detection within X-ray imagery. Although CARAFE \cite{wang19:carafe} is a simpler architecture ($49.41$ million parameters) compared to the Cascade Mask R-CNN \cite{cai19:cascade} ($77.04$ million parameters), it offers a better generalisation ability by training on a more varied set of multiple X-ray imagery variants. In Figure \ref{fig:det_ex1}A the target {\it laptop} is missed in both test images when trained solely on {\it rgb} imagery, but successfully detected when trained with combined \{{\it rgb,hlz}\} X-ray imagery (Figure \ref{fig:det_ex1}B). Hence, we can deduce that although X-ray images are from differing scanners, the transferability of the trained CNN models is significantly improved by training over a more varied training set that includes both pseudo-colour {\it rgb} and variant dual-energy X-ray imagery.
\vspace{-0.2cm}
\newcolumntype{g}{>{\columncolor{blue!5}}c}
\begin{table}[htb!]
\renewcommand*{\arraystretch}{0.8}
\centering
\small
\begin{tabular}{lggg}
\hline
\rowcolor{white}
Training & Model & \multicolumn{2}{c}{Test-set} \\ \cline{3-4}  
\rowcolor{white}
Dataset & & {\it dbs$\_$laptop} & {\it dbr$\_$laptop} \\ \hline
{\it deei6$_{rgb}$} & \small M RCNN  & 0.749 & 0.530 \\
\rowcolor{white}
& \small YOLACT & 0.633 & 0.344 \\
&\small CARAFE &  0.775 & 0.476 \\
\rowcolor{white}
&\small CM RCNN &  0.763 & 0.518 \\ \hline
{\it deei6$_{rgb,hlz}$} & \small M RCNN  & 0.807 & 0.593 \\
\rowcolor{white}
& \small YOLACT  & 0.782 & 0.521 \\
& \small CARAFE & {\bf 0.835} & {\bf 0.611} \\
\rowcolor{white}
& \small CM RCNN & 0.803 & 0.587 \\ \hline
\end{tabular}
\vspace{-0.3cm}
\caption{Object detection results (AP) on {\it dbs$\_$laptop} and {\it dbr$\_$laptop} datasets, where CNN architectures are trained using variant X-ray imagery from the {\it deei6} dataset.}
\label{Table:mAP_generalise}
\end{table}

\vspace{-0.2cm}
    \vspace{-0.8cm}
\section{Conclusion} \label{s:conclusion}
\vspace{-0.4cm}
This work examines the impact of X-ray imagery variants, i.e., dual-energy X-ray responses ({\it high, low}), effective-$z$ and pseudo-colour ({\it rgb}), via the use of CNN architectures for the object detection task posed within X-ray baggage security screening. We illustrate that the combination of {\it rgb, high, low} and effective-$z$ X-ray imagery produces maximal performance across all four CNN architectures for a six classes object detection problem, with CARAFE \cite{wang19:carafe} achieving the highest mAP of $0.7$. Furthermore, our results also demonstrate a remarkable degree of generalisation capability in terms of cross-scanner transferability (AP: $0.835/0.611$ with CARAFE \cite{wang19:carafe}) for a one class object detection problem by combining \{{\it rgb, hlz}\} X-ray imagery. This clearly illustrates a strong insight into the benefits of using a combination of dual-energy X-ray imagery for object detection and segmentation tasks, which could additionally useful for component-wise anomaly detection analysis. Future work will consider the use of dual-energy variant imagery for combined material discrimination and anomaly detection within cluttered X-ray security imagery.

    \small{
\bibliographystyle{IEEEbib}
\bibliography{egbib,object_detection,ref_mod}
}
    
\end{document}